\documentclass[letterpaper, 10 pt, conference]{ieeeconf}  

\IEEEoverridecommandlockouts 

\usepackage{cite}
\usepackage{amsmath,amssymb,amsfonts}
\usepackage{algorithmic}
\usepackage{graphicx}
\usepackage{textcomp}
\usepackage{xcolor}
\usepackage{subcaption}
\usepackage[hidelinks]{hyperref}
\usepackage{url}
\usepackage{booktabs}

\usepackage[markup=bfit, deletedmarkup=sout, authormarkup=brackets]{changes}
\definechangesauthor[name={Filipe}, color=brown]{FG}
\definechangesauthor[name={Valentin}, color=yellow]{VM}
\definechangesauthor[name={Daniela}, color=green]{DC}
\definechangesauthor[name={Matej}, color=orange]{MH}
\definechangesauthor[name={Marianne}, color=magenta]{MJ}

\newcommand{\figref}[1]{Fig.~\ref{#1}}
\newcommand{\tabref}[1]{Table~\ref{#1}}

\newcommand{\eg}{\textit{e.g.},}

\title{\LARGE \bf
Infant simulator with an embodied caregiver: Generating infant-perspective touch and vision during social interaction
}

\author{Noemi Vaculinova and Matej Hoffmann
\thanks{The authors are with Department of Cybernetics, Faculty of Electrical Engineering, Czech Technical University in Prague, Prague, Czech Republic.
{\tt\small matej.hoffmann@fel.cvut.cz}}%
\thanks{This work was supported by the Czech Science Foundation (GA ČR), project no. 25-18113S.}
}

\begin{document}

\maketitle
\thispagestyle{empty}
\pagestyle{empty}

\begin{abstract}
Early development unfolds in caregiver–infant dyads, where infants’ sensorimotor streams are shaped by physical contact and face-to-face interaction. Yet developmental robotics simulators commonly model infants in isolation, limiting the study of caregiver-mediated experience. We present a caregiver-enabled extension of the Multi-Modal Infant Model (MIMo) in MuJoCo that turns MIMo into a controllable platform for replaying dyadic interaction and generating dense infant-perspective observations. The system provides (i) an articulated caregiver model compatible with MIMo morphologies, parameterized from anthropometrics and optionally resized to a recorded caregiver; (ii) a workflow to replay naturalistic caregiver–infant holding and soothing interactions; and (iii) logging and visualizing the infant's first-person multimodal experience (we show touch and vision). We showcase the tool on touch by introducing origin-aware contact logging that disambiguates self-, caregiver-, and environment-generated contact and supports aggregation into touch-rate statistics comparable to manual coding. While we showcase tactile analysis, the platform is intended more broadly as a generator of multimodal dyadic datasets (touch and egocentric vision) for modeling the development of social interaction.

\end{abstract}

\section{INTRODUCTION}
Early development unfolds in caregiver–infant dyads: caregivers shape infants’ sensorimotor statistics through physical handling, affective touch, and close-range face-to-face interaction.
Touch is among the earliest developing senses \cite{myowa2006human} and remains a primary channel for regulation and learning \cite{gallace2010science}. Caregiver touch plays a significant role in early regulation and clinical care (\eg{} soothing touch and stroking)~\cite{puschel2022gentle,madden2025soothing}. In clinical practice, tactile contact such as skin-to-skin (kangaroo care) is associated with improved outcomes, including for preterm infants \cite{Moore2016early,boundy2016kangaroo,who2025}, and affectionate caregiver touch relates to neurodevelopment and dyadic synchrony \cite{carozza2021role}. Behavioral analyses have described major holding positions and their developmental progression~\cite{negayama2010behavioral,romer1990parent}, motivating simulation scenarios in which caregiver posture and contact are central variables.
Similarly, although vision is quite poor at birth, it develops quickly and the details about how often, for how long, and from which angles the caregiver appears in the infant’s visual field may be important to understand the development of face processing \cite{buiatti2019cortical} or the development of social cognition in general \cite{meltzoff2007like,tomasello2005understanding}. 
Addressing these questions requires access to the infant’s first-person experience.

\subsection{Touch and social touch from infant's perspective}
Looking at a single infant, manual video annotation can quantify spontaneous touches of the hand to the body \cite{Thomas2015,dimercurio2018naturalistic,khoury2022self} and touches to the floor~\cite{dimercurio2018naturalistic}, but it is very laborious to code (cf. this dedicated tool \url{https://github.com/ctu-vras/touch-coder}) and does not scale to whole-body contact, occluded body surfaces, or continuous force estimation. 
Pressure mats can capture infant-surface contact only \cite{Kulvicius2023infant}.
Tactile stimuli are processed differently depending on the stimulus identity, namely on whether they are produced by oneself, by another person or by an inanimate object \cite{tame2023emerging}. Thus, the source of contact has important developmental relevance. 
Moreover, for affective touch, there are additional factors such as active generation of the touch or stroking speed. 
A physics-based simulator with the infant and caregiver brings important advantages: whole-body contacts can be recorded and further differentiated into self-touch, caregiver touch, and contacts with the environment.

\subsection{Gaze and sight of the caregiver from infant's perspective}
Seeing the world through the infant's eyes is also critical for our understanding of development. Head-mounted cameras were employed by the groups of Linda Smith and Chen Yu to study what is in toddler's view and the consequences on the development of perception and language \cite{yoshida2008s,smith2015contributions}. There has been less work studying the presence of the caregiver in the infant's view, with notable exceptions \cite{bradshaw2023infant,fausey2016faces,sugden2014spy}. 
However, head-mounted devices may still be intrusive and hard to apply on even younger infants. 
Having both the infant and the caregiver simulated, incl. the visual streams from their eyes, brings several advantages. First, one can rely on infant and caregiver pose estimation from an external camera only and then retarget these motions into the simulator, without the need for wearable devices. The caregiver and infant in the simulator can also move autonomously and any scenario can be simulated. Second, the gaze is composed of the eye direction and head orientation and these can be studied independently in the simulator. In addition, one can also emulate the gradual maturation of infant vision in the simulator, manipulating the field of view, acuity, etc. \cite{lopez2025mimo}. 

\subsection{Baby simulators}
Physics simulation is widely used to generate repeatable, scalable experiments~\cite{collins2021aReview}.
Biomechanical simulators (\eg{} OpenSim~\cite{delp2007opensim}) provide detailed anatomy but can be computationally intensive.
MuJoCo~\cite{todorov2012mujoco} offers fast contact dynamics and is widely used in robot learning research.
Infant-focused simulators include a soft-body fetus-in-uterus model~\cite{kim2022simulating} and MIMo~\cite{mattern2024mimo}, which provides multimodal sensing and a simplified rigid-body infant.
Recent work on MIMo added growth across infant ages and additional sensing features~\cite{lopez2025mimo}.
Motion retargeting approaches can translate captured infant movements into simulations, enabling the estimation of contact streams~\cite{fiala2023baka}.
To our knowledge, existing infant simulators largely lack an embodied caregiver capable of physical interaction and contact generation.

\subsection{Contributions}
Developmental robotics simulators commonly model infants in isolation, limiting the study of caregiver-mediated experience and its developmental consequences.
We address this gap by adding an embodied caregiver to MIMo~\cite{mattern2024mimo}, a MuJoCo-based simulator designed to study early development with multimodal sensing and compatibility with learning pipelines (\eg{} Gym/Gymnasium-style APIs~\cite{brockman2016openaigym,towers2024gymnasium}).
Our goal is not to replace empirical studies but to provide a controllable platform that can generate dense tactile and visual observations from the infant’s perspective during caregiver interaction---including contacts that are difficult to observe in video (\eg{} back-of-head and back-of-torso contact).
This work provides:
\begin{itemize}
  \item \textbf{Caregiver model}: an adult female caregiver compatible with the MIMo ecosystem, parameterized using anthropometric data~\cite{gordon20142012anthropometric} and optionally resized to a specific caregiver.
  \item \textbf{Interaction replay}: a workflow to replay caregiver--infant interactions.
  \item \textbf{Logging and visualizing the infant’s first-person multimodal experience}: We focus on touch and demonstrate also vision from the infant's perspective. We extend MIMo touch processing to separate self- and caregiver-origin touches, log spatiotemporal contact with origin metadata, and generate touch-rate heatmaps over infant body regions.
  \item \textbf{Demonstrations and comparisons}: evaluation of recorded holding/caressing scenarios inspired by behavioral descriptions of infant holding~\cite{negayama2010behavioral,romer1990parent}, replayed with different infant ages~\cite{lopez2025mimo}, and comparison to manually coded infant hand self-touch statistics~\cite{khoury2022self}.
\end{itemize}

\begin{figure*}[htb!]
  \centering
  \includegraphics[width=0.9\linewidth]{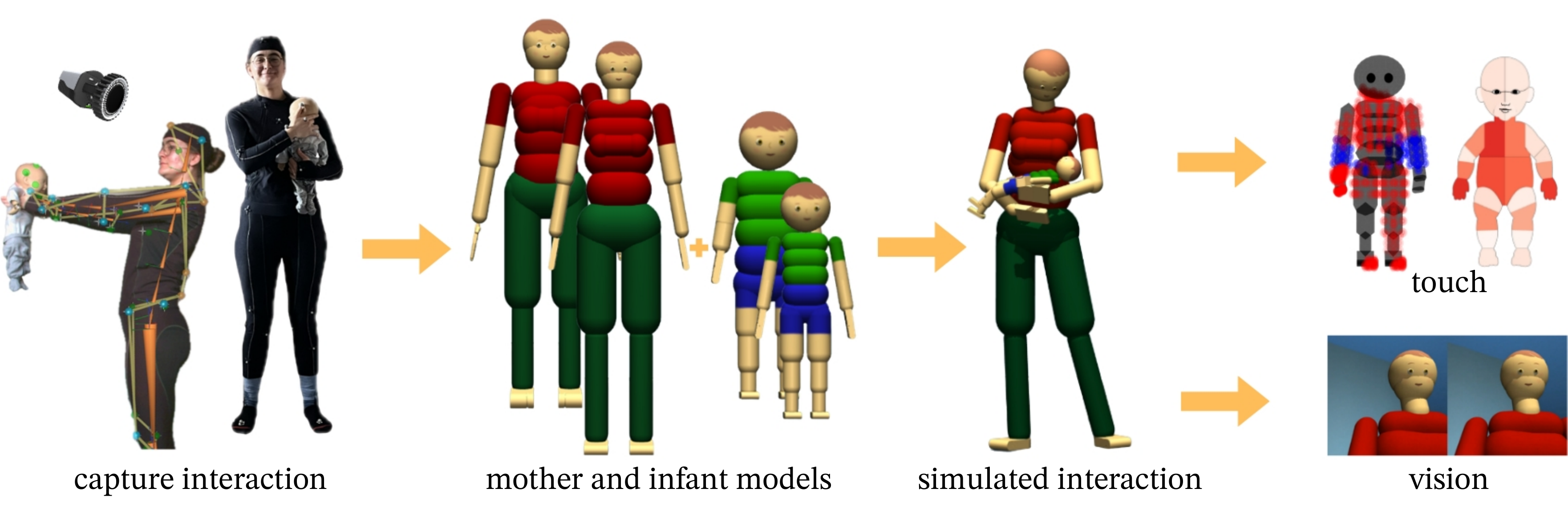}
  \caption{Overview of the reanimation pipeline: the interaction is recorded using optical motion capture, then is applied to previously created caregiver and infant  models, from which we obtain multimodal outputs (touch and vision) with contact logging and heatmap visualization.}
  \label{fig:overview}
\end{figure*}


\section{METHODS}

An overview of the system's functionality is available in the accompanying video.
The code for extending the MIMo simulator with functionalities described in this paper is available in a GitHub repository 
\url{https://github.com/ctu-vras/mom-for-mimo}. 
The recordings of scenarios, processed data, and output videos are in a public Google Drive folder~\cite{vaculinova2025momDrive}. Additional technical details are available in a Master's thesis~\cite{vaculinova2026thesis}. 

\subsection{System overview}

In our expansion of the MIMo simulator, we
(i) model a caregiver compatible with MIMo, (ii) record caregiver--infant interaction motion with an actor holding a newborn-sized doll, (iii) retarget the motion to the simulated caregiver (and partially to the infant) using MuJoCo MoCap bodies, and (iv) log and analyze tactile contact on the infant’s body, distinguishing touch origin (caregiver vs.\ self vs.\ environment).
\figref{fig:overview} summarizes the pipeline leading from the recorded motion to the touch and visual outputs.

\subsection{MIMo infant model and sensing modalities}
MIMo is a MuJoCo-based infant simulator with multimodal sensing, including proprioception, vestibular signals, binocular ``eye'' cameras, and tactile sensing~\cite{mattern2024mimo}.
To support machine learning experiments, MIMo uses a rigid-body and a capsule-based morphology---trading anatomical detail for fast and robust collision handling.
Touch is represented by a set of surface sensors distributed over each body part.
In the newborn-sized model used here, the sensor count is on the order of a few thousand (about three thousand in the default configuration), with denser sampling on hands and feet than on the torso and limbs (mirroring differential tactile acuity).
Sensor density is independent of body size and can be adjusted per body part via modality parameters.

In this work, we focus on two modalities: (i) touch (for analyzing where and by whom the infant is contacted) and (ii) vision (for the infant’s egocentric view during interactions).
We also record a third-person render for qualitative inspection and debugging.

\subsection{Caregiver model}
\begin{figure}
  \centering
  \includegraphics[width=0.85\linewidth]{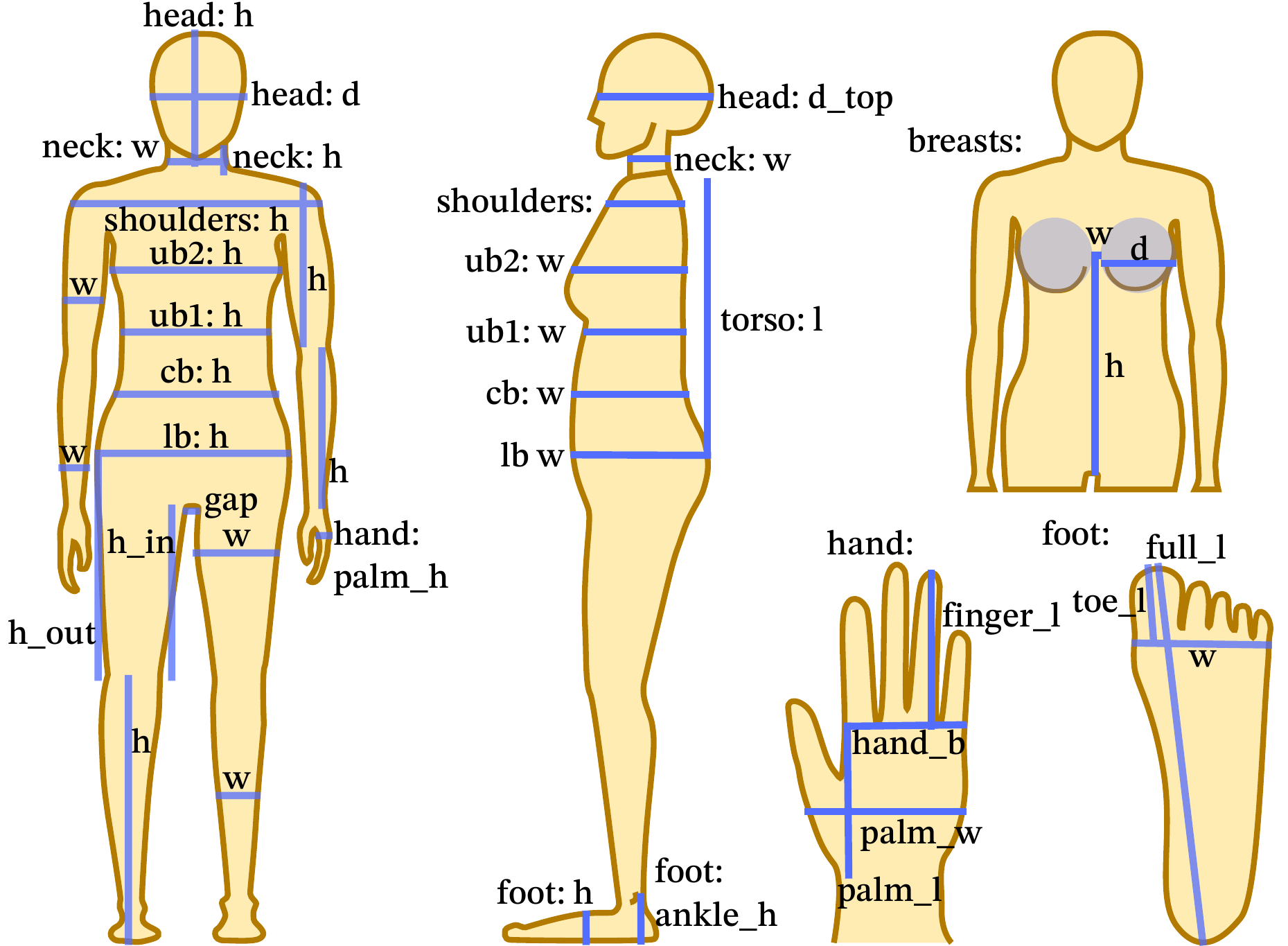}
  \caption{All measurements needed for a custom female caregiver model based on an actor.}
  \label{fig:measure_mom}
\end{figure}

The caregiver is implemented as a rigid-body, capsule-based articulated model, following MIMo’s design philosophy. We created a base model with proportions derived from the 50th percentile female measurements in a large anthropometric study~\cite{gordon20142012anthropometric}.
Segment masses were assigned using established body-segment mass distributions~\cite{plagenhoef1983anatomical} with total body mass set to match the chosen anthropometric percentile. 
To enable subject-specific replay, we provide a resizing procedure that scales the base model to measured limb lengths and circumferences of a recorded caregiver or actor (see \figref{fig:measure_mom}). These measurements are then recalculated to parameters representing the capsules, spheres, boxes, and cylinders. This produces a body model matched to actual recorded caregiver. 

\begin{figure}
  \centering
  \includegraphics[width=0.7\linewidth]{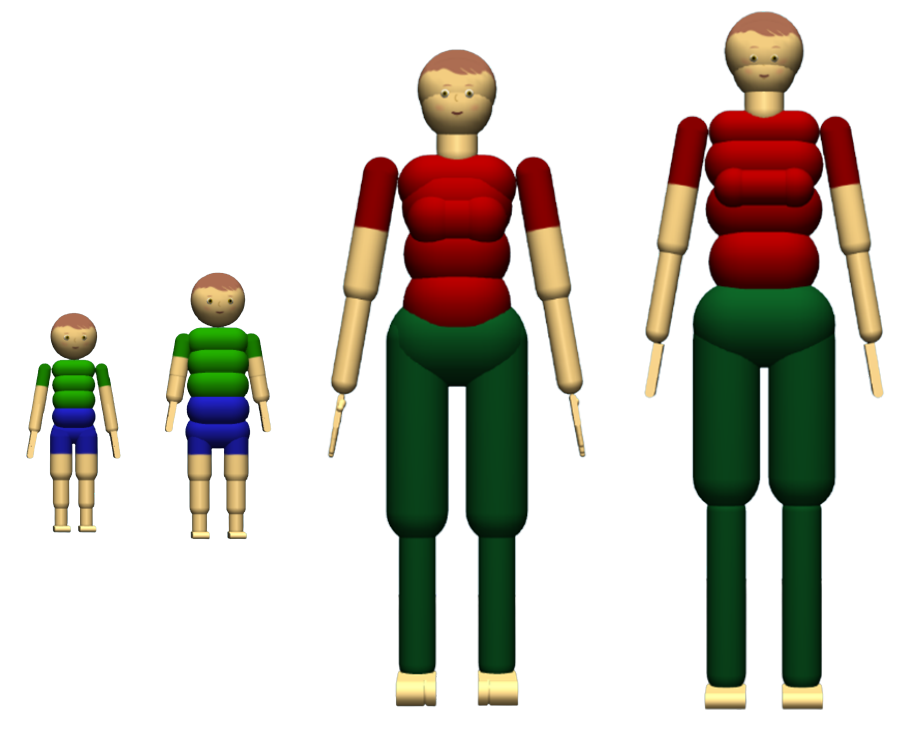}
  \caption{Models used in the experiments (left to right): newborn MIMo, two-month-old MIMo~\cite{lopez2025mimo}, base caregiver, and actor-matched caregiver.}
  \label{fig:models}
\end{figure}

\figref{fig:models} shows the infant and caregiver models used in this paper.
We use the ``simple'' MIMo hand/foot configuration (mitten-like hands) to match the caregiver’s simplified hands and because grasping dexterity is not included in the recorded replay scenarios.

\subsection{Retargeting caregiver motions}\label{sec:mocap_replay}
In this work, we recorded an actor holding a realistic newborn-size stand-in-baby doll in positions inspired by the literature  \cite{negayama2010behavioral,romer1990parent}. The actor was wearing a motion capture suit with markers. However, the pipeline could be modified to use, for example, 3d keypoints extracted from a video recording (see \cite{muller2024generative}).

Using an infrared motion capture camera system, we recorded short mock caregiver--infant interactions (up to 40\,s each) with an actor wearing a marker-based optical motion capture suit with 43 markers while holding and caressing a newborn-sized doll with markers on its head. Then, the recorded marker trajectories were labeled, cleaned, and exported as a time series.

In the simulator, we import the marker trajectories and drive the caregiver using MuJoCo MoCap bodies~\cite{mujocoMocap}. These are defined as massless and shapeless bodies, which are linked to the corresponding bodies of the models through standard MuJoCo equality constraints.
Because the recorded infant doll was not actuated, the simulated infant is treated as a passive physical body, whose pose emerges from physics and caregiver contact, rather than full-body kinematic retargeting.
This design choice trades pose accuracy for physically consistent collision and force generation, which is central to our target outputs (touch and vision). Also, having only three markers on the infant leads to fewer marker occlusions in the captured footage when the actor holds the doll.

A practical issue in close-contact scenarios is camera clipping in the eye views when the infant’s head is pressed against the caregiver body.
In these cases, we reduce the near clipping distance (\texttt{znear}) in the MuJoCo visual parameters to render near-surface geometry, enabling usable egocentric visual streams even in ``hug'' configurations.

\subsection{Touch sensing, origin disambiguation, and time-series logging}\label{sec:timeseries}
MIMo implements touch using several thousand touch sensors distributed over the infant body. 
At each simulation step, MuJoCo provides a list of contacts between bodies. From those, MIMo finds contact pairs with at least one participant equipped with sensors (meaning contacts to the infant). As the bodies used are rigid and simplified, the contact between them culminates in one contact point. To simulate human-like soft touches, the contacts are linearly spread around the contact point for further processing or visualization. The touch output is then a list of activated sensors and the forces acting on them. 

\begin{figure}
    \centering
    \includegraphics[width=0.65\linewidth]{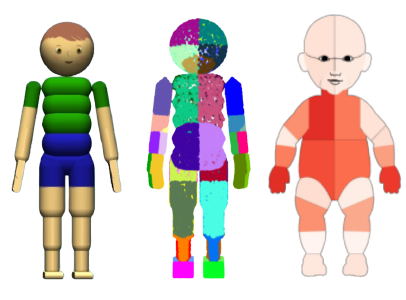}
    \caption{The areas used for touch analysis mapped on the MIMo infant model. Left to right: the MIMo model, the MIMo model segmented into areas by color and the standardized infant figure, which is used in manual annotations~\cite{khoury2022self}.}
    \label{fig:areas}
\end{figure}

We extend MIMo’s touch processing to
assign each contact a \textit{touch origin category} as a context indicator about the participant who touches the infant sensors (\textit{infant} for self-touches, \textit{caregiver}, or \textit{environment}). 
For touch analysis purposes, we further map sensors to a coarser set of 50 body regions (front and back of the head, torso, upper/lower limbs, hands, feet, etc.), consistent with prior touch statistics heatmap visualizations~\cite{khoury2022self}; see \figref{fig:areas} for a visualization of the areas on MIMo and the standardized infant illustration.

\section{EXPERIMENTS AND RESULTS}
The experiments were designed to demonstrate the functionality of the tool, i.e. an infant and caregiver in a physics-based simulator. We paid specific attention to the processing and logging of touch on the part of the infant. The first-person visual streams are available, but will not be analyzed in detail.
The interactions are mock, but inspired by common holding and soothing poses. If real interaction motion data---poses of caregiver and infant---were available, the same pipeline can be applied to generate first-person sensory streams during such interaction. 

\subsection{Interaction scenarios}

\begin{figure*}
  \centering
  \subfloat[Hold close (baby\_arms1).]{\includegraphics[width=0.3\textwidth]{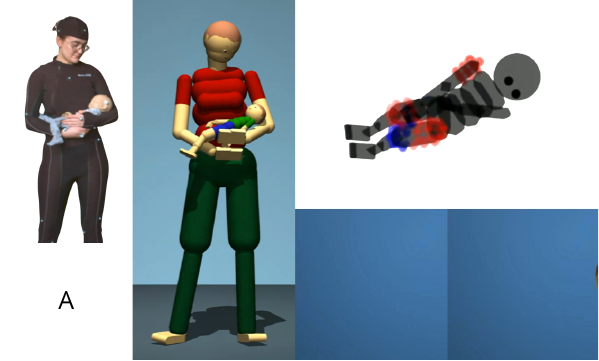}\label{fig:scenA}}
  \hfil
  \subfloat[Hold away (baby\_away2).]{\includegraphics[width=0.31\textwidth]{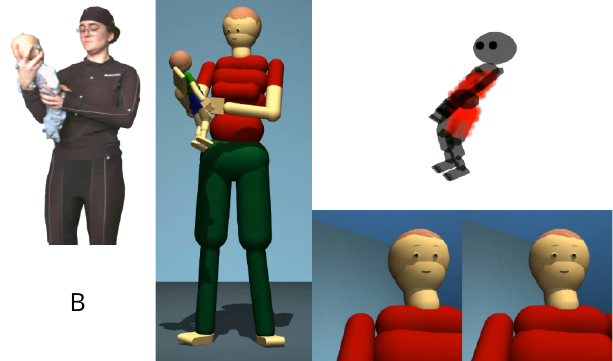}\label{fig:scenB}}
  \hfil
  \subfloat[Caress (touch1).]{\includegraphics[width=0.3\textwidth]{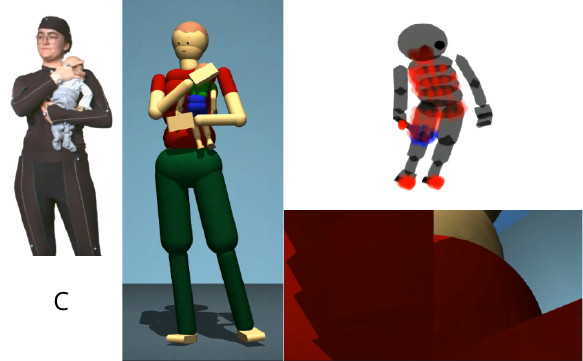}\label{fig:scenC}}
  \caption{Representative scenarios. Each panel shows (left-to-right within the panel): reference video frame, third person simulation render, touch visualization (caregiver-origin in red; self-origin in blue), and infant left/right eye views.}
  \label{fig:scenarios}
\end{figure*}

\begin{figure}
    \centering
    \includegraphics[width=0.85\linewidth]{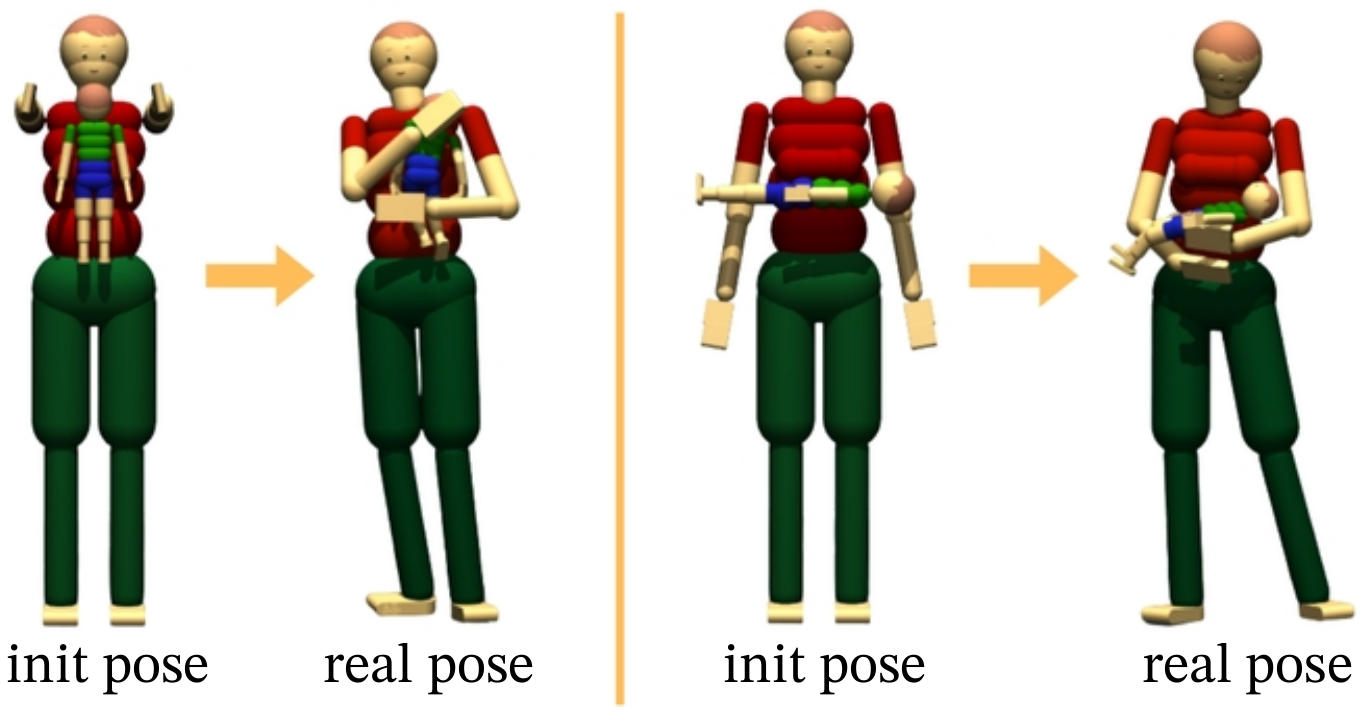}
    \caption{The two initialization poses of the caregiver and child with an example of a real pose to be adapted. Left is the pose for a vertical child, whereas the right is for the child in a horizontal position.}
    \label{fig:init_pose}
\end{figure}

We recorded eight interaction scenarios inspired by common holding and soothing poses~\cite{negayama2010behavioral, romer1990parent}. These were then processed and replayed as Section \ref{sec:mocap_replay} describes.
\tabref{tab:recordings} lists the scenarios recorded for the experiments. We will describe three scenarios that showcase different types of poses struck by the female actor and the newborn-sized doll (see \figref{fig:scenarios}). The other scenarios are variations of the three chosen scenarios.
One of two initialization poses (depending on the desired poses to be adopted by the caregiver and infant, see \figref{fig:init_pose}) is used to ensure the simulated bodies settle into a configuration compatible with the incoming marker trajectories before replay begins; this reduces ``snagging'' contacts at simulation start and improves tracking.

\begin{table}[t]
\caption{Recorded caregiver--infant interaction scenarios used for simulation replay. The interactions introduced in \ref{sec:replay and out} are in bold.}
\label{tab:recordings}
\centering
\scriptsize
\begin{tabular}{@{}l l r@{}}
\toprule
\textbf{Scenario} & \textbf{Description} & \textbf{Frames} \\\midrule
\textbf{baby\_arms1} & infant held close; caregiver sways & 4000\\
baby\_arms2 & variant of baby\_arms1 & 3978\\
baby\_arms4 & infant lifted to chest (pose change) & 1466\\
\textbf{baby\_away2} & infant held away; face-to-face view & 2742\\
baby\_shoulder1 & infant supported on shoulder & 3000\\
baby\_shoulder2 & variant of shoulder hold & 1671\\
\textbf{touch1} & infant held to chest; back caressing & 1458\\
touch2 & variant of touch1 & 3999\\
\bottomrule
\end{tabular}
\end{table}

\subsection{Qualitative replay and multimodal outputs}\label{sec:replay and out}
For each replay, we generate: (i) a third-person scene render, (ii) infant eye-camera views, and (iii) touch logging and visualization classified by the source that colors activated infant touch sensors by the origin of the touch (caregiver in red; self in blue; other in yellow).
\figref{fig:scenarios} and the accompanying video illustrate the three representative interaction scenarios from \tabref{tab:recordings} that will be described in the following sections.

Scenario A (\textit{baby\_arms1}) shows a close hold with substantial caregiver contact on the infant’s torso and back as the caregiver sways.
Scenario B (\textit{baby\_away2}) is ``face-to-face''. Infant eye views consistently contain the caregiver head/upper body, illustrating the platform’s ability to generate first-person visual exposure statistics. Self-touch is reduced because the infant is lightly supported by the caregiver’s hands.
Scenario C (\textit{touch1}) combines a chest hold with localized caressing on the back; here, the touch-origin-aware visualization highlights caregiver-originated contact on the back and side of the torso.


\subsection{Validation of the touch outputs}
\begin{figure}
    \centering
    \includegraphics[width=\linewidth]{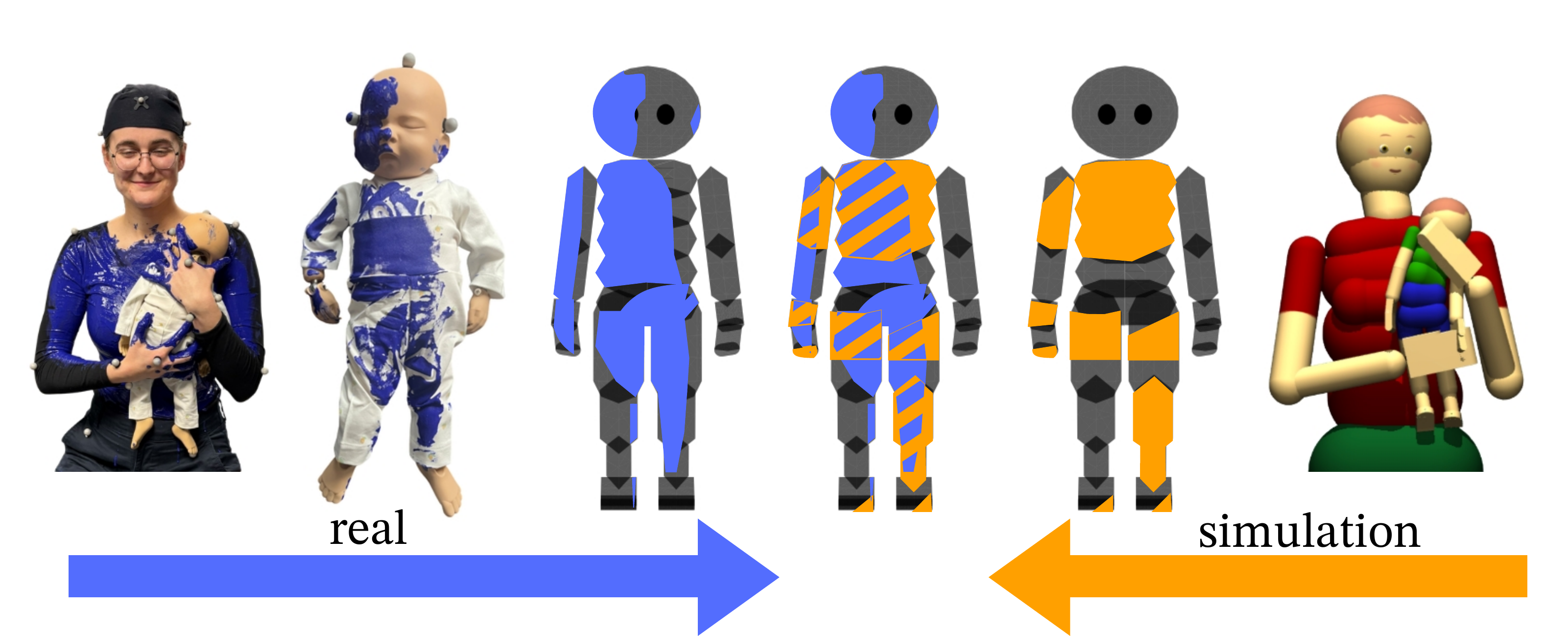}
    \caption{The comparison of the real (blue) and simulated (orange) touches and their overlap.}
    \label{fig:validation}
\end{figure}

To provide a frame of reference for comparison with the simulated touch, we recorded three static poses (similar to the ones from \figref{fig:scenarios}) of the adult actor and the newborn doll. The actor's upper body was covered in fresh, skin-safe paint, which then transferred to the infant's body in points of contact. The colored infant body was then photographed and digitized (see \figref{fig:validation}). 
Then, we compared these touches to those from the simulated touch sensors on the infant body.

Most touched areas were detected in the simulation, with the specific placements varying in accuracy, as can be seen in \figref{fig:validation}. The main limitation is the lack of simulated touches to the head, which suffers most from the shape simplification, as a sphere does not properly represent the complex shape. Also, the use of motion capture markers on the head does not allow the head to settle into the caregiver's arms.
Furthermore, the MIMo touch spreading, which simulates soft touches, is not perfect, as there is only one point of contact when two rigid bodies meet. Whereas the real touches happen over the entire touched area and are influenced by the originator.
Overall, the simulated touch information is valuable if we keep the limitations in mind.



\begin{figure*}
  \centering
  \subfloat[Scenario A: baby\_arms1.]{\includegraphics[width=0.325\textwidth]{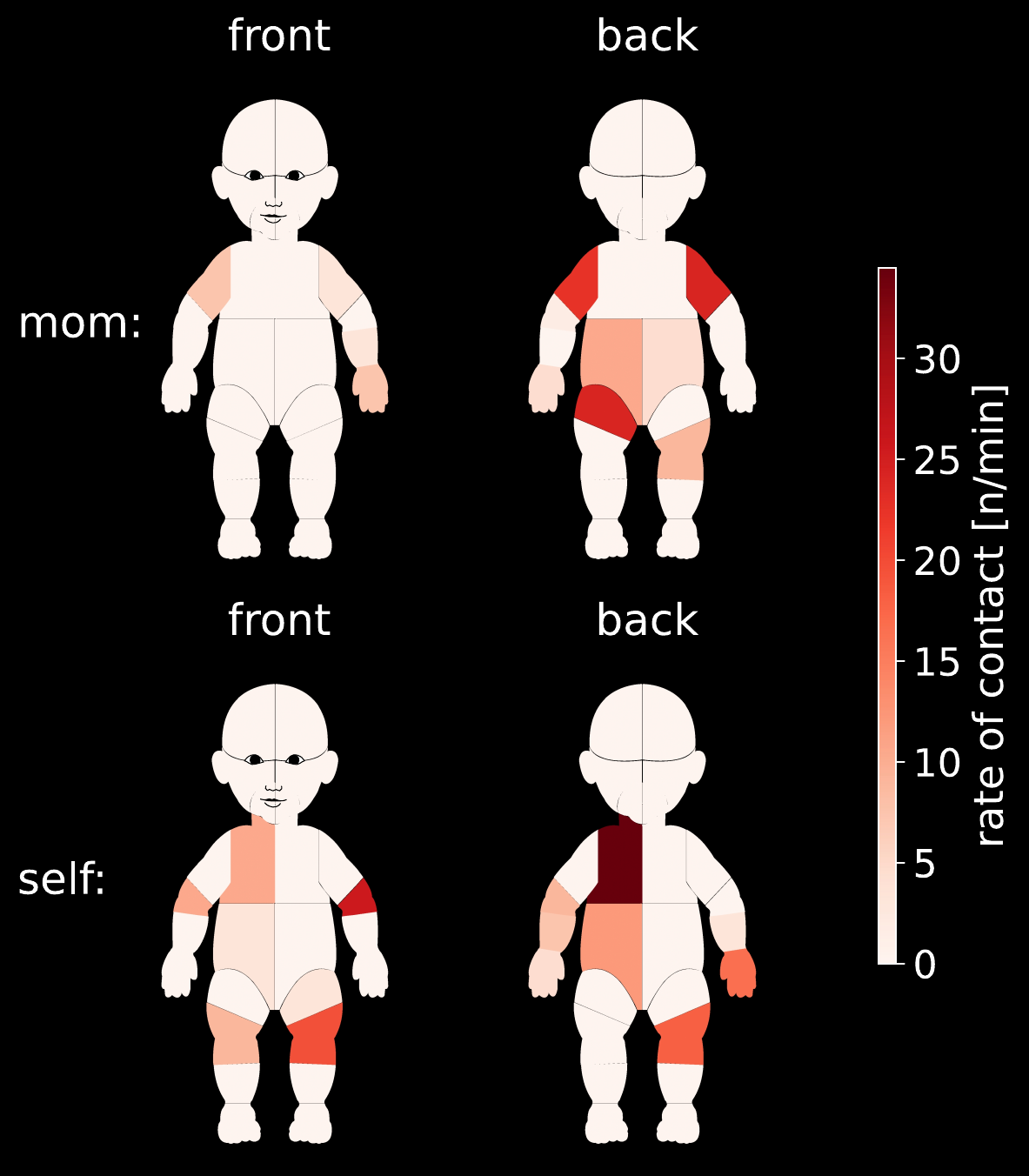}\label{fig:hmA}}
  \hfil
  \subfloat[Scenario B: baby\_away2.]{\includegraphics[width=0.325\textwidth]{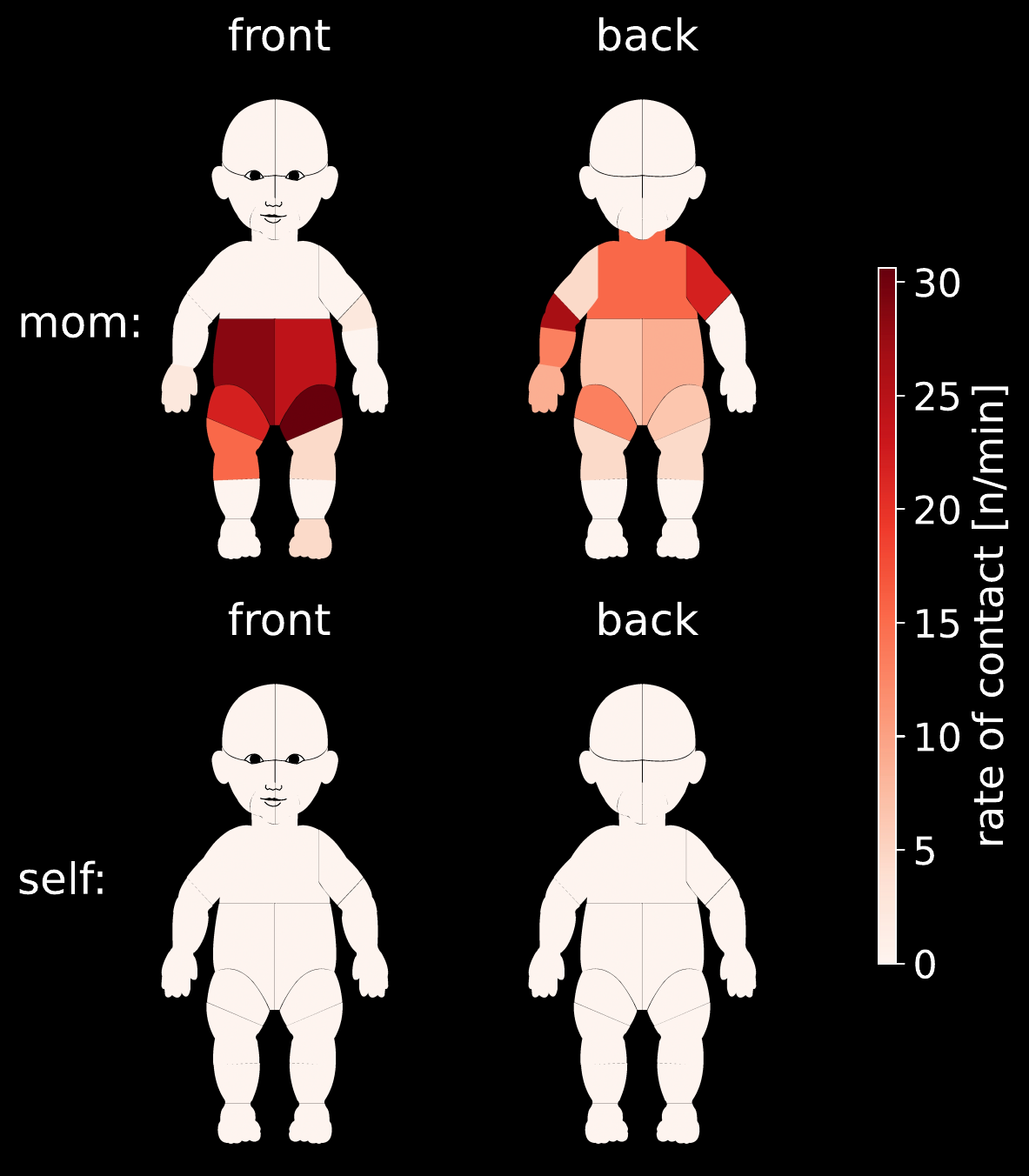}\label{fig:hmB}}
  \hfil
  \subfloat[Scenario C: touch1.]{\includegraphics[width=0.325\textwidth]{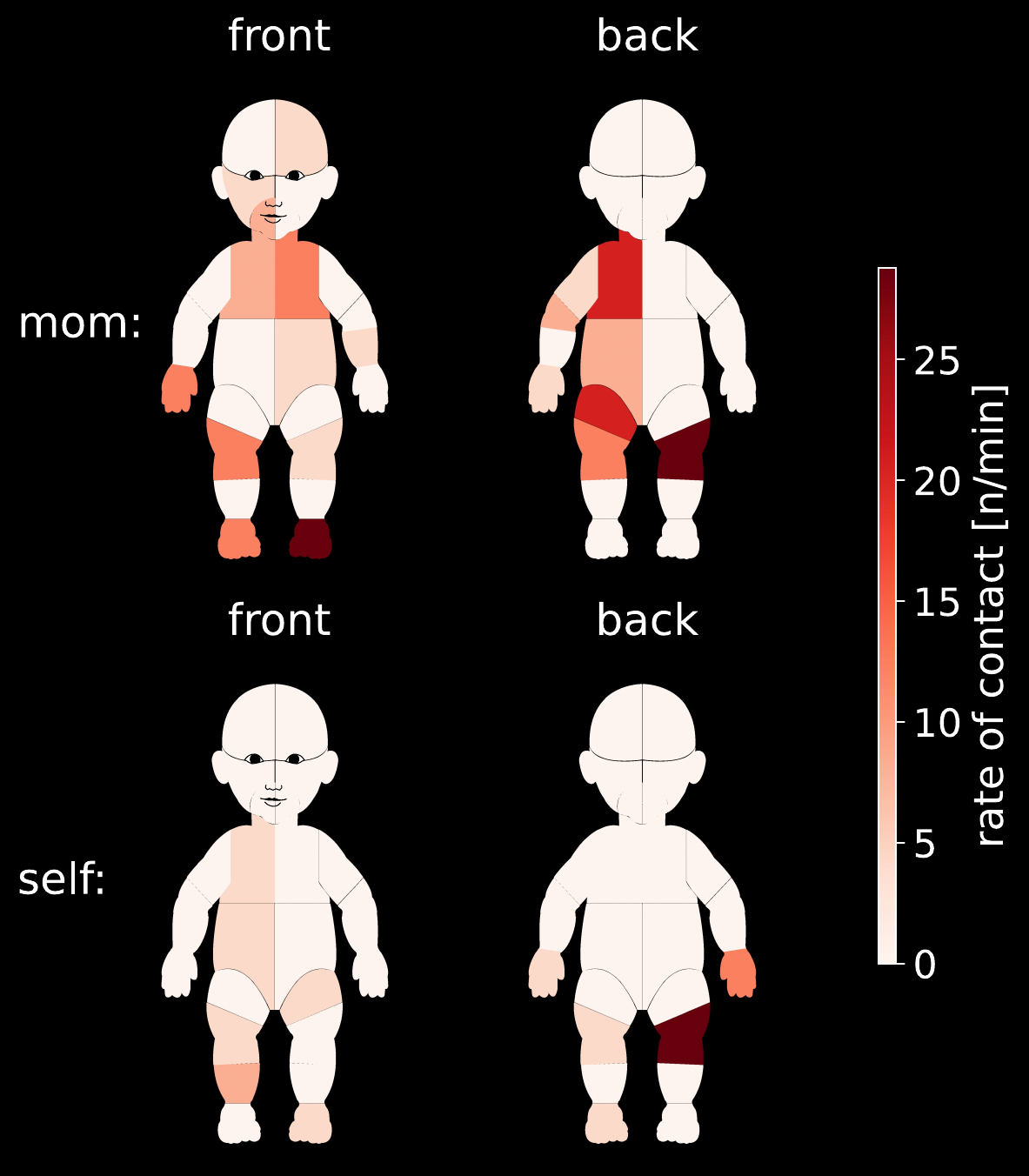}\label{fig:hmC}}
  \caption{Scenario-level touch-rate heatmaps (touches/min). For each scenario, caregiver-origin and self-origin contacts are shown on front/back body regions.}
  \label{fig:scenario_heatmaps}
\end{figure*}

\subsection{Scenario-level touch patterns}
Using the logged time series (from Section \ref{sec:timeseries}), we compute touch-rate heatmaps (in touches/min) for each scenario by aggregating touch events
per body region and normalizing them by replay duration. 
\figref{fig:scenario_heatmaps} shows heatmaps for the three scenarios in \figref{fig:scenarios}.
The close-hold scenario (\figref{fig:hmA}) concentrates caregiver contact on the back and torso, while the away-hold (\figref{fig:hmB}) shows more localized support at the shoulders and trunk.
In the caressing scenario (\figref{fig:hmC}), elevated touch rates appear on the back regions corresponding to the caregiver’s stroking motion.
Self-touch in these scenarios is largely induced by caregiver motion and body compression (\eg{} limbs contacting torso during swaying), consistent with the passive infant model.

\subsection{Replaying the same interactions with different infant ages}
\begin{figure}
  \centering
  \subfloat[Newborn.]{\includegraphics[width=0.49\linewidth]{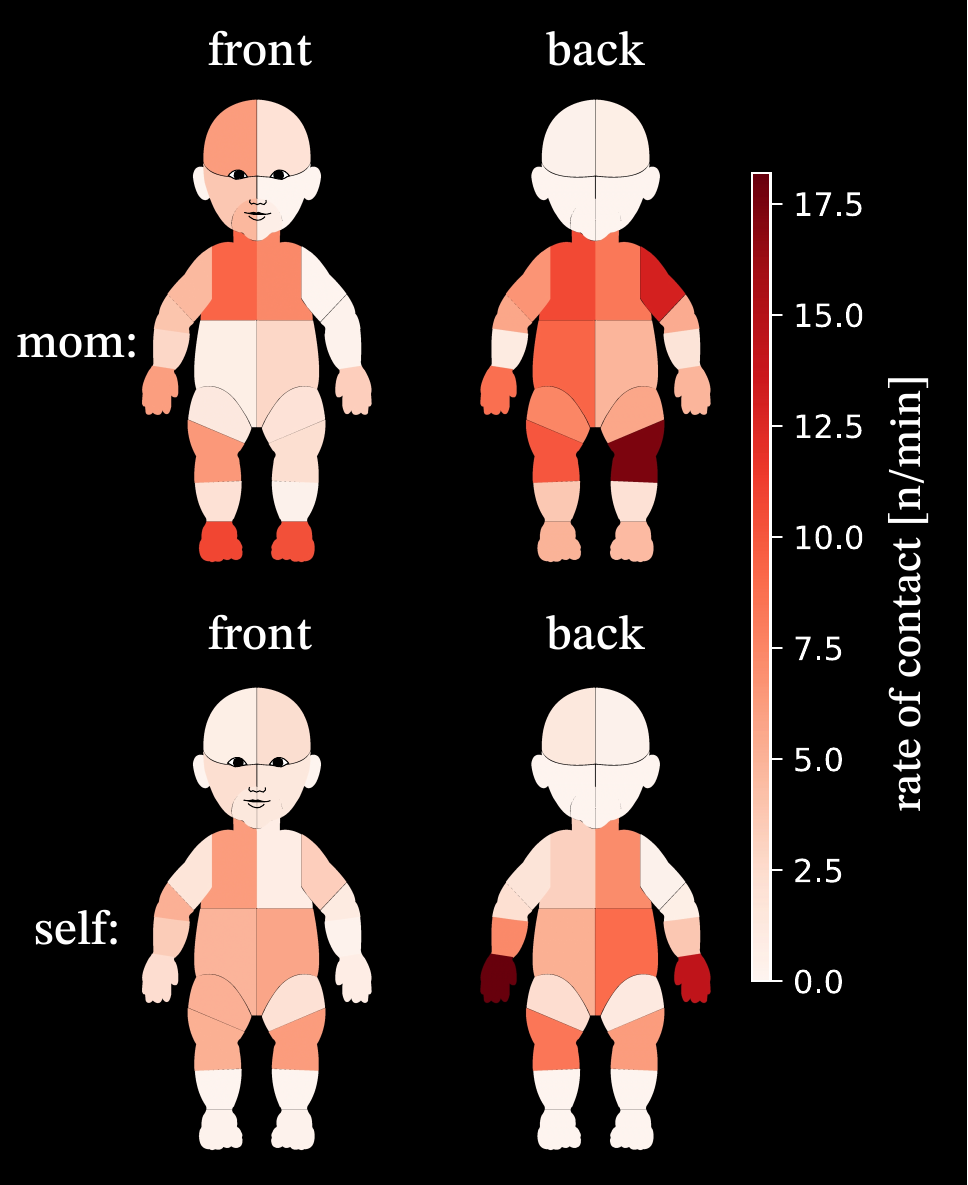}\label{fig:age_nb}}
  \hfil
  \subfloat[Two-month-old.]{\includegraphics[width=0.477\linewidth]{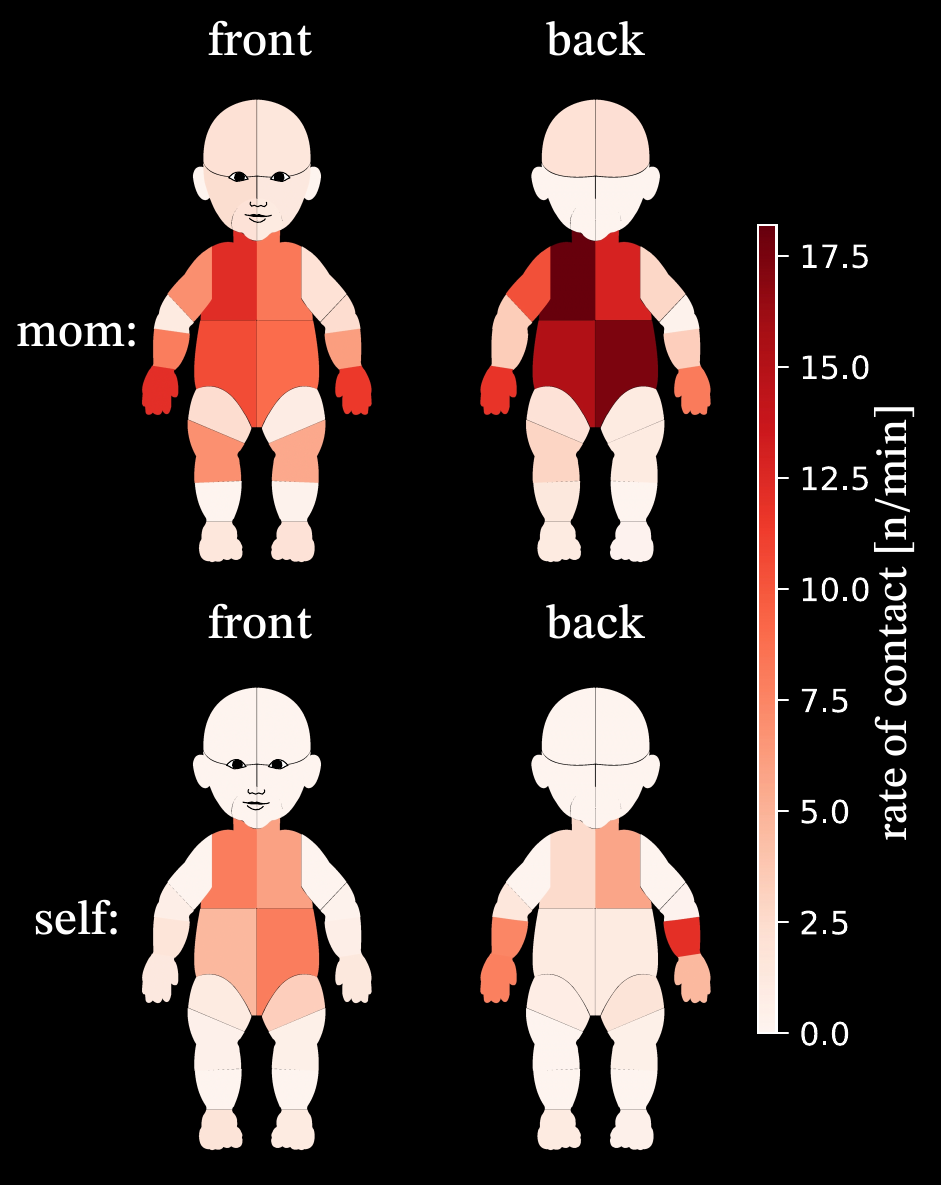}\label{fig:age_2mo}}
  \caption{Aggregated touch-rate heatmaps (touches/min) over all replayed scenarios for two infant ages. Back and front surfaces are shown separately within each heatmap.}
  \label{fig:ageheat}
\end{figure}

To demonstrate compatibility with MIMo’s growth mechanism~\cite{lopez2025mimo}, we replayed each scenario with a newborn and a two-month-old infant model (keeping the caregiver motion and initial conditions fixed).
\figref{fig:ageheat} shows aggregated touch-rate heatmaps over all replays from \tabref{tab:recordings} for both ages.
The two-month model exhibits a more torso-dominated caregiver contact distribution and reduced leg contact in several holds, consistent with a larger body that does not ``nest'' in the caregiver’s arms in the same way as a newborn.
This illustrates how a single caregiver motion dataset can be reused to probe developmental changes in contact statistics induced purely by morphology.

\subsection{Comparison to manual self-touch statistics}
\begin{figure}
    \centering
    \includegraphics[width=\linewidth]{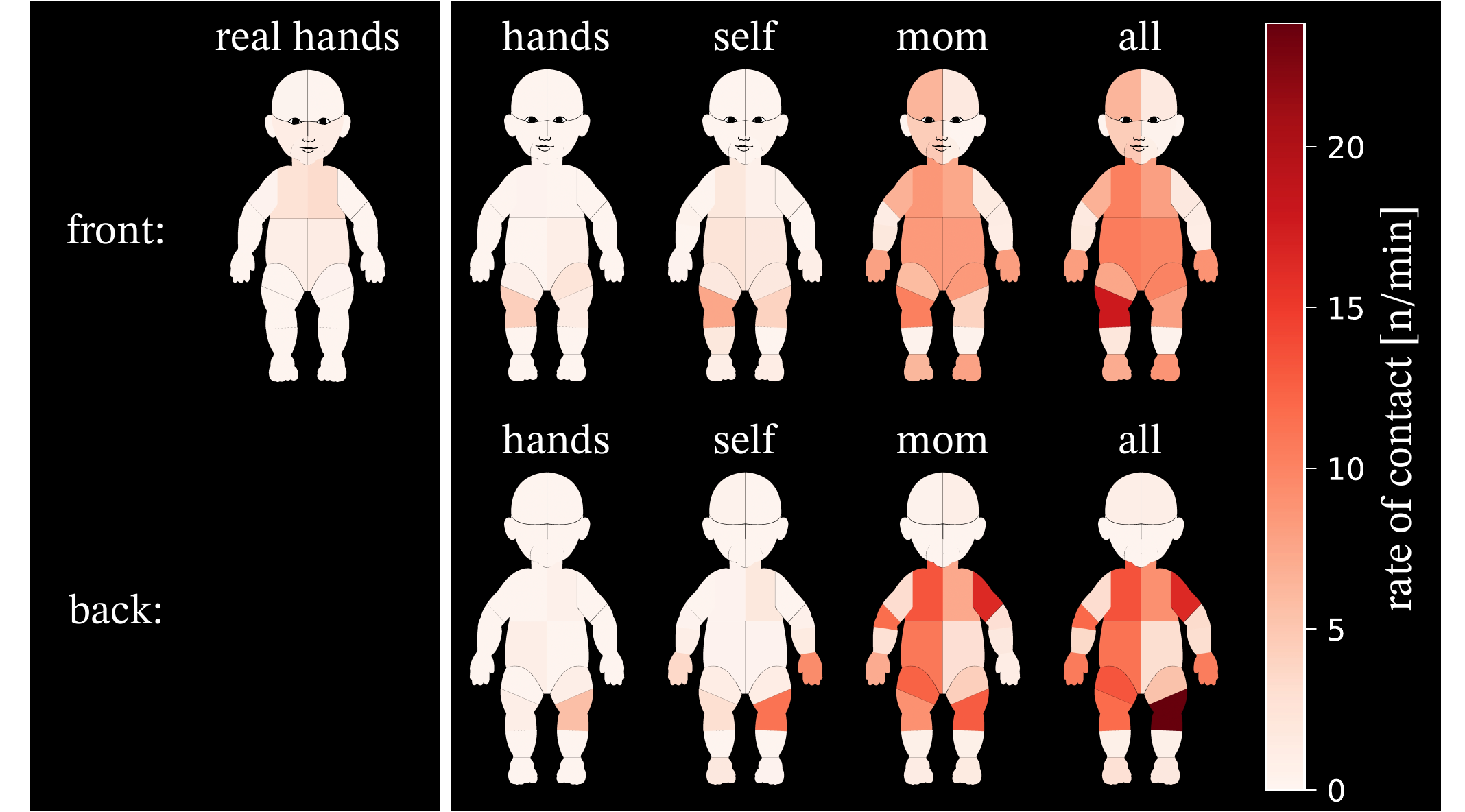}
    \caption{Comparison of touch statistics: manually coded real infant hand self-touch~\cite{khoury2022self} vs.\ touch distributions obtainable from the simulator by filtering contact origin (hands, self, caregiver, all).}
    \label{fig:jasno_validation}
\end{figure}

A practical advantage of simulation is the short processing time and dense, origin-aware contact logging, including contacts to body regions occluded in videos.
To contextualize our outputs, we compare simulated touch distributions to manually annotated self-touch statistics from real infant videos~\cite{khoury2022self}. These videos depict infant self-touch with hands in a supine position. 
\figref{fig:jasno_validation} shows the touch rate (touches/min) for: real hand self-touch (video-coded), simulated hand self-touch (isolated by origin), simulated whole-body self-contact, simulated caregiver-origin contact, and all contacts combined.

The simulated \textit{hand} self-touch rates are broadly comparable to the magnitude of real infant hand self-touch rates, while the simulation additionally provides back-of-body contact not accessible from typical video views.
The caregiver-origin category naturally dominates the total contact rate in many replayed holding and soothing actions, highlighting a key difference from isolated infant recordings.
Our simulated hand self-touch is biased toward body regions that are incidentally compressed during caregiver holding; this differs from real infants, who actively self-explore.
This mismatch is expected given our passive infant and emphasizes the need to integrate active infant control for developmental hypotheses focused on endogenous exploration.



\section{DISCUSSION AND FUTURE WORK}
A key motivation for simulation is repeatability.
Once a caregiver motion sequence is captured and retargeted, it can be replayed deterministically under varied infant morphology, sensor parameters, or experimental conditions (\eg{} different caregiver stiffness, alternative contact kernels).
The origin-aware contact representation enables analyses beyond heatmaps, including temporal structure of touch bouts, contact durations, and sequences of caregiver vs.\ self-contact.

The present system is intended as a first step toward embodied computational studies of caregiver-infant multimodal interaction.
In the following, we outline concrete research directions enabled by the caregiver extension and changes required to make the simulator suitable for more mechanistic developmental hypotheses.

The tool needs to be validated regarding the fidelity of the emulated sensorimotor experiences. We performed preliminary validation of the fidelity of the contact generation using painted T-shirts. For the visual sensory emulation, the baby doll or infants carried by a caregiver could wear a head-mounted camera to allow comparison with the view from the simulator after retargeting. 

\subsection{Infant and caregiver marker-less pose estimation from video}
There is rapid development of computer vision techniques for marker-less human pose estimation from images or videos. In \cite{diaz2024estimation}, 2D pose estimation (i.e. in the image) using OpenPose and depth from the Kinect sensor were combined for pose estimation in mother-infant interaction.  
This is a challenging task since deep learning based methods are trained on adult datasets, creating a domain shift for infants (but see \cite{gama2025automatic} for a comparison of 2D pose estimators on infant datasets). 
Moreover, close interaction of two persons creates specific challenges as the identified body keypoints overlap (see e.g., \cite{purkrabek2025detection}). 
To capture the full richness of the infant-caregiver interaction, 3D shape and pose estimation are likely required. Many tools are available for 3D adult pose and shape estimation (e.g., \cite{SMPL-X:2019,bregier2025human}). An infant rather than adult shape is desired (see the SMIL model \cite{hesse2019learning} as the infant variant of the SMPL model). 
The reconstructed poses and shapes can then theoretically be applied to automatic contact labeling, but typical algorithms for 3D pose estimation (e.g., SMPLify-x \cite{SMPL-X:2019}) have priors to prevent self-collisions. The work of Muller et al.~\cite{muller2023self,muller2024generative} constitutes an exception targeting self-and interpersonal contact in 3D human mesh reconstruction.
A promising direction could be: (1) Perform the pose estimation from video and extract movements of the caregiver and infant; (2) Use motion retargeting to map them onto the models of the caregiver and infant in MuJoCo (see \cite{fiala2023baka,lopez2026simulating} for an infant retargeting variant); (3) generate the multimodal experiences of the infant and the caregiver.   
A professionally captured dataset, the BONES-SEED~\cite{bones-seed}, can be used as a source of caregiver motion, as there are multiple scenarios involving a baby. However, the current open-source scenarios do not include the infant and its motion.

\subsection{Pose fidelity vs.\ physically consistent contact}
Because our target output is tactile contact from the infant’s perspective, we prioritized physically consistent collision and force generation over the exact replication of infant limb kinematics.
Sparse infant constraints (limited markers) mean the simulated infant pose can differ from the recorded doll, especially in leg and head support.
Future work should incorporate richer infant motion capture or markerless pose estimation to constrain the infant more tightly without sacrificing contact realism.

\subsection{Higher-fidelity contact and perception}
The caregiver and infant are simplified rigid-body models, which improves simulation speed but can miss contacts that would occur with deformable soft tissue (\eg{} head-arm cradling or chest compression).
Similarly, touch sensors approximate distributed skin receptors via surface sampling and a contact-spreading kernel; while useful for analysis and learning, this is not a substitute for high-fidelity tactile skin models.
Soft-body simulation~\cite{kim2022simulating} or layered ``visual'' meshes over collision capsules could improve realism.

For questions where tissue compliance matters (hugging, chest-to-chest skin-to-skin), future versions should incorporate deformable elements or soft contact models, similar in spirit to fetal-in-uterus simulation~\cite{kim2022simulating}.
A practical compromise is to keep collision proxies capsule-based (for robustness) but render a visually realistic ``skin'' mesh; this would improve the ecological validity of infant egocentric vision (\eg{} for studying how caregivers occupy the field of view~\cite{franchak2024developmental}) while preserving stable contact dynamics.
More detailed hands (articulated fingers) would also enable modeling of grasp-like supports and fine caressing patterns that are currently lost with mitten-like hands.


\subsection{From motion replay to interactive dyads}
In this paper, the movements of the caregiver were extracted using motion capture and the infant was entirely passive, with motions induced only by the caregiver. Active infant motions could be added: reactive or with more complex internal dynamics. The caregiver movements could also be autonomously generated. Eventually, both the caregiver and the infant could autonomously generate behaviors, in a closed-loop fashion taking into account the behavior of the interaction partner.   
Such interactive dyads could be used to examine hypotheses about the developmental role of caregiver touch~\cite{carozza2021role,boundy2016kangaroo} and to generate synthetic multimodal datasets (touch and egocentric vision) for perception and learning models. 
Since MIMo already targets learning-compatible interfaces~\cite{brockman2016openaigym,towers2024gymnasium}, the caregiver extension can be used to train policies that optimize goals defined on multimodal sensory histories (\eg{} maintaining face visibility while minimizing uncomfortable contact forces).



\subsection{Using simulation outputs in empirical pipelines}
Beyond hypothesis generation, the simulator can support empirical work by providing synthetic annotations that are difficult to obtain in video: dense contact labels (who touched whom, where, and with what force), occluded-body contact, and egocentric visual streams.
These outputs can be used to (i) train and validate vision models that infer contact or holding posture from monocular video, (ii) design experimental protocols that maximize observability of target variables, and (iii) construct priors for inverse problems that reconstruct infant sensorimotor experience from sparse observations~\cite{kanazawa2022open-ended}.
In this sense, the caregiver-enabled MIMo platform can act as a bridge between computational modeling and measurement-limited behavioral datasets.

\section{CONCLUSION}
We presented a first step toward simulating caregiver-infant physical interaction within a developmental robotics platform by adding a physics-based caregiver model to MIMo.
The system supports motion-capture-driven replay of naturalistic holding and caressing and extends MIMo’s touch sensing to log and analyze contact by origin.
Demonstrations show that the same caregiver motion can be replayed across infant ages and that simulation enables contact statistics that are difficult to obtain empirically (\eg{} back-of-body and caregiver-origin contact).
The resulting platform opens a path toward interactive dyadic simulations that couple touch and vision from the infant's perspective and can be used to generate hypotheses and datasets for early development research.

\section*{ACKNOWLEDGMENT}
We thank the MIMo developers for making the simulator available~\cite{mattern2024mimo}. Mainly, we extend our gratitude to Francisco M. López for valuable consultations about the MIMo simulator. We also thank Jason Khoury, who supported the motion capture setup, touch-analysis tooling and provided us with the manually coded data.


\bibliographystyle{./bibliography/IEEEtran}
\bibliography{./bibliography/references}

@misc{bones-seed,
    author = {},
    title = {{BONES-SEED ‑ Skeletal Everyday Embodied Dataset}},
    url = {https://bones.studio/datasets/seed},
    publisher = {Bones Studio},
    year = {2026},
  note = {{Accessed: 2026-3-24}}
}

@article{boundy2016kangaroo,
  title={{Kangaroo Mother Care and Neonatal Outcomes: A Meta-analysis}},
  author={Boundy, Ellen O and Dastjerdi, Roya and Spiegelman, Donna and Fawzi, Wafaie W and Missmer, Stacey A and Lieberman, Ellice and Kajeepeta, Sandhya and Wall, Stephen and Chan, Grace J},
  journal={Pediatrics},
  volume={137},
  number={1},
  year={2016},
  publisher={American Academy of Pediatrics}
}

@article{bradshaw2023infant,
  title={Infant embodied attention in context: Feasibility of home-based head-mounted eye tracking in early infancy},
  author={Bradshaw, Jessica and Fu, Xiaoxue and Yurkovic-Harding, Julia and Abney, Drew},
  journal={Developmental Cognitive Neuroscience},
  volume={64},
  pages={101299},
  year={2023},
  publisher={Elsevier}
}

@article{bregier2025human,
  title={Human Mesh Modeling for Anny Body},
  author={Br{\'e}gier, Romain and Fiche, Gu{\'e}nol{\'e} and Bravo-S{\'a}nchez, Laura and Lucas, Thomas and Armando, Matthieu and Weinzaepfel, Philippe and Rogez, Gr{\'e}gory and Baradel, Fabien},
  journal={arXiv preprint arXiv:2511.03589},
  year={2025}
}

@article{brockman2016openaigym,
  title={{OpenAI Gym}},
  author={Brockman, Greg and Cheung, Vicki and Pettersson, Ludwig and Schneider, Jonas and Schulman, John and Tang, Jie and Zaremba, Wojciech},
  journal={arXiv preprint arXiv:1606.01540},
  year={2016}
}

@article{buiatti2019cortical,
  title={Cortical route for facelike pattern processing in human newborns},
  author={Buiatti, Marco and Di Giorgio, Elisa and Piazza, Manuela and Polloni, Carlo and Menna, Giuseppe and Taddei, Fabrizio and Baldo, Ermanno and Vallortigara, Giorgio},
  journal={Proceedings of the National Academy of Sciences},
  volume={116},
  number={10},
  pages={4625--4630},
  year={2019},
  publisher={National Academy of Sciences}
}

@article{carozza2021role,
  title={{The Role of Affectionate Caregiver Touch in Early Neurodevelopment and Parent–Infant Interactional Synchrony}},
  author={Carozza, Sofia and Leong, Victoria},
  journal={Frontiers in Neuroscience},
  volume={14},
  year={2021},
  publisher={Frontiers Media SA}
}

@ARTICLE{collins2021aReview,
  author={Collins, Jack and Chand, Shelvin and Vanderkop, Anthony and Howard, David},
  journal={IEEE Access}, 
  title={{A Review of Physics Simulators for Robotic Applications}}, 
  year={2021},
  volume={9},
  number={},
  pages={51416-51431}
}

@ARTICLE{delp2007opensim,
  author={Delp, Scott L. and Anderson, Frank C. and Arnold, Allison S. and Loan, Peter and Habib, Ayman and John, Chand T. and Guendelman, Eran and Thelen, Darryl G.},
  journal={IEEE Transactions on Biomedical Engineering}, 
  title={{OpenSim: Open-Source Software to Create and Analyze Dynamic Simulations of Movement}}, 
  year={2007},
  volume={54},
  number={11},
  pages={1940-1950},
  doi={10.1109/TBME.2007.901024}
}

@article{diaz2024estimation,
  title={Estimation of human body 3D pose for parent-infant interaction settings using azure Kinect and OpenPose},
  author={Diaz-Rojas, Fran{\c{c}}oise and Myowa, Masako},
  journal={MethodsX},
  volume={13},
  pages={102861},
  year={2024},
  publisher={Elsevier}
}

@article{dimercurio2018naturalistic,
  title={{A Naturalistic Observation of Spontaneous Touches to the Body and Environment in the First 2 Months of Life}},
  author={DiMercurio, Abigail and Connell, John P and Clark, Matthew and Corbetta, Daniela},
  journal={Frontiers in Psychology},
  volume={9},
  year={2018},
  publisher={Frontiers Media SA}
}

@article{fausey2016faces,
  title={From faces to hands: Changing visual input in the first two years},
  author={Fausey, Caitlin M and Jayaraman, Swapnaa and Smith, Linda B},
  journal={Cognition},
  volume={152},
  pages={101--107},
  year={2016},
  publisher={Elsevier}
}

@misc{fiala2023baka,
  author = {Fiala, Ondrej},
  title = {{Retargeting Infant Movements to Baby Humanoid Robots}},
  year = {2023},
  note = {{Czech Technical University (CTU), Bachelor's Thesis}}
}

@article{franchak2024developmental,
  title={{Developmental Changes in How Head Orientation Structures Infants’ Visual Attention}},
  author={Franchak, John M and Smith, Linda and Yu, Chen},
  journal={Developmental psychobiology},
  volume={66},
  number={7},
  year={2024},
  publisher={Wiley Online Library}
}

@article{gallace2010science,
  title={{The science of interpersonal touch: An overview}},
  author={Gallace, Alberto and Spence, Charles},
  journal={Neuroscience \& Biobehavioral Reviews},
  volume={34},
  number={2},
  pages={246--259},
  year={2010},
  publisher={Elsevier}
}

@article{gama2025automatic,
  title={{Automatic infant 2D pose estimation from videos: comparing seven deep neural network methods}},
  author={Gama, Filipe and M{\'\i}sa{\v{r}}, Mat{\v{e}}j and Navara, Luk{\'a}{\v{s}} and T. Popescu, Sergiu and Hoffmann, Matej},
  journal={Behavior Research Methods},
  volume={57},
  number={10},
  pages={280},
  year={2025},
  publisher={Springer}
}

@article{gordon20142012anthropometric,
  title={{2012 Anthropometric Survey of U.S. Army Personnel: Methods and Summary Statistics}},
  author={Gordon, Claire C and Blackwell, Cynthia L and Bradtmiller, Bruce and Parham, Joseph L and Barrientos, Patricia and Paquette, Stephen P and Corner, Brian D and Carson, Jeremy M and Venezia, Joseph C and Rockwell, Belva M and others},
  journal={Army Natick Soldier Research Development and Engineering Center MA, Tech. Rep},
  year={2014}
}

@article{hesse2019learning,
  title={{Learning and Tracking the 3D Body Shape of Freely Moving Infants from RGB-D sequences}},
  author={Hesse, Nikolas and Pujades, Sergi and Black, Michael J and Arens, Michael and Hofmann, Ulrich G and Schroeder, A Sebastian},
  journal={IEEE Transactions on Pattern Analysis and Machine Intelligence},
  volume={42},
  number={10},
  pages={2540--2551},
  year={2019},
  publisher={IEEE}
}

@article{kanazawa2022open-ended,
author = {Kanazawa, Hoshinori and Yamada, Yasunori and Tanaka, Kazutoshi and Kawai, Masahiko and Niwa, Fusako and Iwanaga, Kogoro and Kuniyoshi, Yasuo},
year = {2022},
month = {12},
pages = {},
title = {{Open-ended movements structure sensorimotor information in early human development}},
volume = {120},
journal = {Proceedings of the National Academy of Sciences},
doi = {10.1073/pnas.2209953120}
}

@inproceedings{khoury2022self,
  title={{Self-touch and other spontaneous behavior patterns in early infancy}},
  author={Khoury, Jason and Popescu, Sergiu T and Gama, Filipe and Marcel, Valentin and Hoffmann, Matej},
  booktitle={2022 IEEE International Conference on Development and Learning (ICDL)},
  pages={148--155},
  year={2022},
  organization={IEEE}
}

@inproceedings{kim2022simulating,
  title={{Simulating a Human Fetus in Soft Uterus}},
  author={Kim, Dongmin and Kanazawa, Hoshinori and Kuniyoshi, Yasuo},
  booktitle={2022 IEEE International Conference on Development and Learning (ICDL)},
  pages={135--141},
  year={2022},
  organization={IEEE}
}

@article{Kulvicius2023infant,
   title={{Infant movement classification through pressure distribution analysis}},
   volume={3},
   ISSN={2730-664X},
   DOI={10.1038/s43856-023-00342-5},
   number={1},
   journal={Communications Medicine},
   publisher={Springer Science and Business Media LLC},
   author={Kulvicius, Tomas and Zhang, Dajie and Nielsen-Saines, Karin and Bölte, Sven and Kraft, Marc and Einspieler, Christa and Poustka, Luise and Wörgötter, Florentin and Marschik, Peter B.},
   year={2023},
   month=aug }

@inproceedings{lopez2025mimo,
  title={{MIMo Grows! Simulating Body and Sensory Development in a Multimodal Infant Model}},
  author={L{\'o}pez, Francisco M and Lenz, Miles and Fedozzi, Marco G and Aubret, Arthur and Triesch, Jochen},
  booktitle={2025 IEEE International Conference on Development and Learning (ICDL)},  
  year={2025},
  organization={IEEE}
}

@InProceedings{lopez2026simulating,
  author    = {Lopez, F.M. and Kanazawa, H. and Fiala, O. and Balashov, Y. and Marcel, V. and Rustler, L. and Lenz, M. and Kim, D. and Kuniyoshi, Y. and Triesch, J. and Hoffmann, M.},
  booktitle = {IEEE International Conference on Development and Learning (ICDL)},
  title     = {Simulating Infant First-Person Sensorimotor Experience via Motion Retargeting from Babies to Humanoids},
  year      = {2026},
}

@article{madden2025soothing,
  title={{Soothing touch matters: Patterns of everyday mother-infant physical contact and their real-time physiological implications}},
  author={Madden-Rusnak, Anna and Micheletti, Megan and Bailey, Loryn and de Barbaro, Kaya},
  journal={Infant Behavior and Development},
  volume={78},
  year={2025},
  publisher={Elsevier}
}

@article{mattern2024mimo,
  title={{MIMo: A Multi-Modal Infant Model for Studying Cognitive Development}},
  author={Mattern, Dominik and Schumacher, Pierre and L{\'o}pez, Francisco M and Raabe, Marcel C and Ernst, Markus R and Aubret, Arthur and Triesch, Jochen},
  journal={IEEE Transactions on Cognitive and Developmental Systems},
  year={2024},
  publisher={IEEE}
}

@article{Moore2016early,
    author = {Moore, E R and Bergman, N and Anderson, G C and Medley, N},
    title = {{Early skin‐to‐skin contact for mothers and their healthy newborn infants}},
    journal = {Cochrane Database of Systematic Reviews},
    number = {11},
    year = {2016},
    publisher = {John Wiley & Sons, Ltd}
}

@article{meltzoff2007like,
  title={‘Like me’: a foundation for social cognition},
  author={Meltzoff, Andrew N},
  journal={Developmental science},
  volume={10},
  number={1},
  pages={126--134},
  year={2007},
  publisher={Wiley Online Library}
}

@online{mujocoMocap, 
  title={{Modelling - MuJoCo documentation: MoCap bodies}}, 
  url={https://mujoco.readthedocs.io/en/stable/modeling.html#mocap-bodies},
  journal={MuJoCo documentation},
  note = {Accessed: 2025-12-15} }

@inproceedings{muller2024generative,
  title={Generative proxemics: A prior for 3d social interaction from images},
  author={M{\"u}ller, Lea and Ye, Vickie and Pavlakos, Georgios and Black, Michael and Kanazawa, Angjoo},
  booktitle={Proceedings of the IEEE/CVF Conference on Computer Vision and Pattern Recognition},
  pages={9687--9697},
  year={2024}
}

@phdthesis{muller2023self,
  title={{Self-and Interpersonal Contact in 3D Human Mesh Reconstruction}},
  author={M{\"u}ller, Lea},
  year={2023},
  school={Dissertation, T{\"u}bingen, Universit{\"a}t T{\"u}bingen}
}

@article{myowa2006human,
  title={{Do Human Fetuses Anticipate Self-Oriented Actions? A Study by Four-Dimensional (4D) Ultrasonography}},
  author={Myowa-Yamakoshi, Masako and Takeshita, Hideko},
  journal={Infancy},
  volume={10},
  number={3},
  pages={289--301},
  year={2006},
  publisher={Taylor \& Francis}
}

@article{negayama2010behavioral,
  title={{Behavioral development of infant holding and its laterality in relation to mothers’ handedness and child-care attitude}},
  author={Negayama, K and Kawai, M and Yamamoto, H and Tomiwa, K and Sakakihara, Y and Japan Children's Study Group and others},
  journal={Infant Behavior and Development},
  volume={33},
  number={1},
  pages={68--78},
  year={2010},
  publisher={Elsevier}
}

@article{plagenhoef1983anatomical,
  title={{Anatomical Data for Analyzing Human Motion}},
  author={Plagenhoef, Stanley and Evans, F Gaynor and Abdelnour, Thomas},
  journal={Research Quarterly for Exercise and Sport},
  volume={54},
  number={2},
  pages={169--178},
  year={1983},
  publisher={Taylor \& Francis}
}

@inproceedings{purkrabek2025detection,
  title={{Detection, Pose Estimation and Segmentation for Multiple Bodies: Closing the Virtuous Circle}},
  author={Purkrabek, Miroslav and Matas, Jiri},
  booktitle={Proceedings of the IEEE/CVF International Conference on Computer Vision},
  pages={9004--9013},
  year={2025}
}

@article{puschel2022gentle,
  title={{Gentle as a mother's touch: C-tactile touch promotes autonomic regulation in preterm infants}},
  author={P{\"u}schel, Isabella and Reichert, J{\"o}rg and Friedrich, Yvonne and Bergander, J{\"o}rg and Weidner, Kerstin and Croy, Ilona},
  journal={Physiology \& Behavior},
  volume={257},
  year={2022},
  publisher={Elsevier}
}

@article{romer1990parent,
  title={{Parent-Infant Holding Patterns and Their Impact on Infant Perceptual and Interactional Experience}},
  author={Romer, Georg and Sossin, K Mark},
  journal={Pre- and Peri-Natal Psychology},
  volume={5},
  pages={69--85},
  year={1990}
}

@article{smith2015contributions,
  title={Contributions of head-mounted cameras to studying the visual environments of infants and young children},
  author={Smith, Linda B and Yu, Chen and Yoshida, Hanako and Fausey, Caitlin M},
  journal={Journal of cognition and development},
  volume={16},
  number={3},
  pages={407--419},
  year={2015},
  publisher={Taylor \& Francis}
}

@inproceedings{SMPL-X:2019,
title = {Expressive Body Capture: 3D Hands, Face, and Body from a Single Image},
author = {Pavlakos, Georgios and Choutas, Vasileios and Ghorbani, Nima and Bolkart, Timo and Osman, Ahmed A. A. and Tzionas, Dimitrios and Black, Michael J.},
booktitle = {Proceedings IEEE Conf. on Computer Vision and Pattern Recognition (CVPR)},
year = {2019}
}

@article{sugden2014spy,
  title={I spy with my little eye: Typical, daily exposure to faces documented from a first-person infant perspective},
  author={Sugden, Nicole A and Mohamed-Ali, Marwan I and Moulson, Margaret C},
  journal={Developmental psychobiology},
  volume={56},
  number={2},
  pages={249--261},
  year={2014},
  publisher={Wiley Online Library}
}

@article{tame2023emerging,
  title={Emerging principles in functional representations of touch},
  author={Tam{\`e}, Luigi and Longo, Matthew R},
  journal={Nature Reviews Psychology},
  volume={2},
  number={8},
  pages={459--471},
  year={2023},
  publisher={Nature Publishing Group US New York}
}

@Article{Thomas2015,
  Title                    = {Independent development of the Reach and the Grasp in spontaneous self-touching by human infants in the first 6 months},
  Author                   = {Thomas, Brittany L and Karl, Jenni M and Whishaw, Ian Q},
  Journal                  = {Frontiers in psychology},
  Year                     = {2015},
  Pages                    = {1526},
  Volume                   = {5},
  Publisher                = {Frontiers}
}

@inproceedings{todorov2012mujoco,
  title={{MuJoCo: A physics engine for model-based control}},
  author={Todorov, Emanuel and Erez, Tom and Tassa, Yuval},
  booktitle={2012 IEEE/RSJ International Conference on Intelligent Robots and Systems},
  pages={5026--5033},
  year={2012},
  organization={IEEE}
}

@article{tomasello2005understanding,
  title={Understanding and sharing intentions: The origins of cultural cognition},
  author={Tomasello, Michael and Carpenter, Malinda and Call, Josep and Behne, Tanya and Moll, Henrike},
  journal={Behavioral and brain sciences},
  volume={28},
  number={5},
  pages={675--691},
  year={2005},
  publisher={Cambridge University Press}
}

@inproceedings{towers2024gymnasium,
      title={{Gymnasium: A Standard Interface for Reinforcement Learning Environments}}, 
      author={Mark Towers and Ariel Kwiatkowski and Jordan Terry and John U. Balis and Gianluca De Cola and Tristan Deleu and Manuel Goulão and Andreas Kallinteris and Markus Krimmel and Arjun K G and Rodrigo Perez-Vicente and Andrea Pierré and Sander Schulhoff and Jun Jet Tai and Hannah Tan and Omar G. Younis},
      year={2025},
      eprint={2407.17032},
      archivePrefix={arXiv},
      primaryClass={cs.LG},
      volume={39},
      booktitle = {Conference on Neural Information Processing Systems (NeurIPS 2025)}
}

@misc{vaculinova2025momDrive,
  author = {Vaculínová, Noemi},
  title = {{Mom for MIMo data}},
  year = {2025},
  publisher = {Google Drive},
  URL = {https://drive.google.com/drive/folders/1AQJQcR1kb0gsvE73_v49ukzLqOPVfkB7?usp=sharing}
}

@misc{vaculinova2026thesis,
  author = {Vaculínová, Noemi},
  title = {{Adding a Caregiver to a Baby Simulator}},
  year = {2026},
  note = {{Czech Technical University (CTU), Master's Thesis}}
}

@book{who2025,
  author    = {{World Health Organization}},
  title     = {{Kangaroo Mother Care: A Clinical Practice Guide}},
  year      = {2025},
  address   = {Geneva},
  publisher = {World Health Organization}
}

@article{yoshida2008s,
  title={What's in view for toddlers? Using a head camera to study visual experience},
  author={Yoshida, Hanako and Smith, Linda B},
  journal={Infancy},
  volume={13},
  number={3},
  pages={229--248},
  year={2008},
  publisher={Wiley Online Library}
}

\end{document}